\documentclass[10pt,twocolumn,letterpaper]{article}

\usepackage[pagenumbers]{cvpr} 

\definecolor{cvprblue}{rgb}{0.21,0.49,0.74}
\usepackage[pagebackref,breaklinks,colorlinks,allcolors=cvprblue]{hyperref}
\usepackage{multirow} 
\usepackage{tikz}
\usepackage{tabularx,ragged2e}
\usetikzlibrary{decorations.pathreplacing}
\usepackage{float}
\usepackage{array}
\usepackage{etoc}
\def\paperID{11649} 
\def\confName{CVPR}
\def\confYear{2025}

\title{MobileH2R: Learning Generalizable Human to Mobile Robot Handover Exclusively from Scalable and Diverse Synthetic Data}

\author{
Zifan Wang\textsuperscript{\textasteriskcentered1,2}~~
Ziqing Chen\textsuperscript{\textasteriskcentered1,2}~~
Junyu Chen\textsuperscript{\textasteriskcentered1}~~
Jilong Wang\textsuperscript{2,3}~~
Yuxin Yang\textsuperscript{2}\\
Yunze Liu\textsuperscript{1,5}~~
Xueyi Liu\textsuperscript{1,5}~~
He Wang\textsuperscript{2,3}~~
Li Yi\textsuperscript{\textdagger1,4,5}
\smallskip\\
\textsuperscript{1}Tsinghua University~~
\textsuperscript{2}Galbot~~
\textsuperscript{3}Peking University~\\
\textsuperscript{4}Shanghai Artificial Intelligence Laboratory~~
\textsuperscript{5}Shanghai Qi Zhi Institute
\\
\href{https://Mobile.github.io}{\textcolor{magenta}{https://MobileH2R.github.io}}
}

\begin{document}

\twocolumn[{%
\maketitle
\vspace{-14pt}
\begin{figure}[H]
\hsize=\textwidth 
\centering
\includegraphics[width=2\columnwidth]{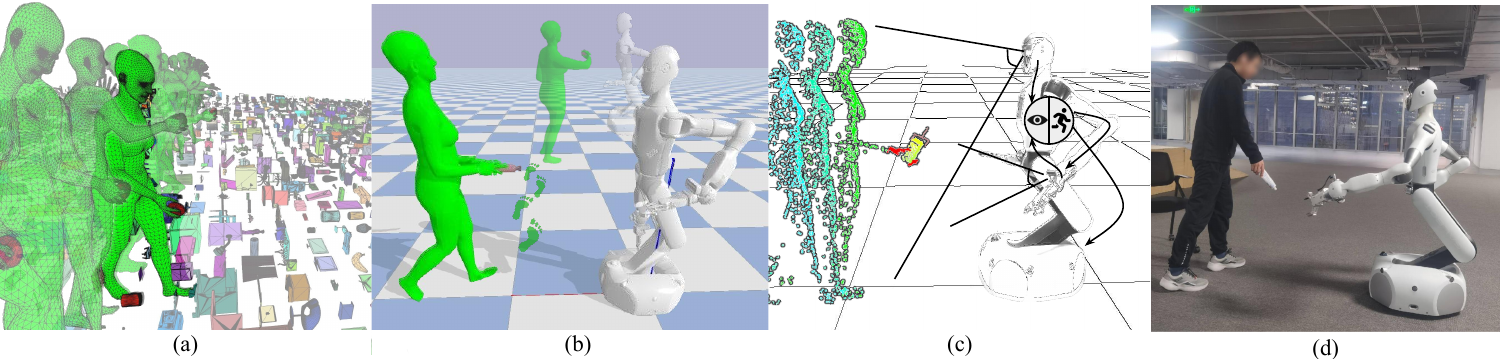}
\vspace{-4pt}
\caption{
\textbf{The overview of MobileH2R.} We propose a framework for generalizable human-to-mobile-robot handover, including a scalable pipeline for diverse full-body human motion synthesis (a), an automatic method for producing safe, imitation-friendly demonstrations (b), an efficient 4D imitation learning approach to learn coordinated base-arm actions (c), and successful sim2real transfer (d).}
\label{fig:intro}
\end{figure}
}]

\begin{abstract}
This paper introduces MobileH2R, a framework for learning generalizable vision-based human-to-mobile-robot (H2MR) handover skills. Unlike traditional fixed-base handovers, this task requires a mobile robot to reliably receive objects in a large workspace enabled by its mobility. Our key insight is that generalizable handover skills can be developed in simulators using high-quality synthetic data, without the need for real-world demonstrations. To achieve this, we propose a scalable pipeline for generating diverse synthetic full-body human motion data, an automated method for creating safe and imitation-friendly demonstrations, and an efficient 4D imitation learning method for distilling large-scale demonstrations into closed-loop policies with base-arm coordination. Experimental evaluations in both simulators and the real world show significant improvements (at least +15\% success rate) over baseline methods in all cases. Experiments also validate that large-scale and diverse synthetic data greatly enhances robot learning, highlighting our scalable framework.
\end{abstract}

\section{Introduction}
\label{sec:intro}







The embodied AI research community has long been motivated by the goal of enabling robots to interact and collaborate naturally with humans. A key challenge is the development of human-to-mobile-robot (H2MR) handover, which requires robots to reliably receive objects handed over by humans across a large, reachable space enabled by their mobility. This capability is crucial in diverse settings such as healthcare assistants and industrial assembly lines, where mobile robots must seamlessly interpret and respond to human actions to perform tasks efficiently and safely.

However, learning generalizable interaction skills between humans and robots presents several unique challenges. Unlike human-free robot manipulation, human-to-robot (H2R) handover remains difficult to scale effectively. Real-world H2R training introduces significant safety risks and costs, making direct training with humans inherently non-scalable. Simulated environments have been proposed as a solution to circumvent these challenges, but creating large-scale, realistic human motion and object assets is itself a demanding task. The first H2R simulation HandoverSim~\cite{chao2022handoversim} uses mocap data DexYCB~\cite{chao2021dexycb} in simulation realistic assets in simulation, but with only 1,000 sequences, it lacks the complexity needed to reflect real-world interactions.

The challenges intensify in H2MR handover, where robots must interpret full-body human actions while managing mobility and navigation complexities. Traditional approaches break down handover into subtasks like grasp estimation and trajectory planning~\cite{yang2021reactive,10608385}, which limits holistic environmental modeling and hinders generalizability. Recent methods, such as GenH2R~\cite{Wang_2024_CVPR}, advance with end-to-end frameworks that scale through synthetic assets, but these often constrain interactions to fixed-base robotic arms or only capture partial human motions, like hand gestures, without full-body movement modeling.

In this work, we aim to achieve generalizable H2MR handover by addressing these challenges. Our key insight is that generalizable handover skills can be developed in simulators without the need for real-world demonstrations, as long as high-quality synthetic data—comprising human motion assets, object assets, and robot demonstrations—is utilized.
We introduce a comprehensive solution that scales both assets and demonstrations, enabling effective closed-loop visuomotor policy learning through imitation.

To scale up synthetic and diverse full-body human motion data, we introduce an automatic pipeline emphasizing both diversity and fidelity. Existing datasets like AMASS~\cite{AMASS:ICCV:2019}, and MotionX~\cite{lin2024motionxlargescale3dexpressive} lack task-specific specialization as well as interactive behaviors for complex HRI tasks like handover. Therefore, We propose a two-stage pipeline for full-body handover motion synthesis. By leveraging some generic motion synthesis algorithms~\cite{karunratanakul2023guidedmotiondiffusioncontrollable}, we generate a wide range of realistic full-body motions, while a task-specific synthesis method creates diverse hand and arm movements tailored for handovers. Furthermore, we design an interactive human agent to respond to the robot’s proximity. Our synthetic dataset includes over 100K interactive handover scenes, scalable for task-specific HRI training.

To ensure safe interaction, we opt to learn interaction policies through imitation rather than reinforcement learning. 
Inspired by GenH2R~\cite{Wang_2024_CVPR}, we explore motion planning methods for demonstration generation. 
In terms of safety, we optimize the planner to ensure that the planned trajectories avoid collisions with the human body and prevent entering the human's blind side. To further facilitate the imitation from the demonstrations, we also ensure that the vision sequence paired with the planned trajectory provides clear and informative object state estimates. Since object states are strongly correlated with the robot's actions, this also strengthens the connection between the vision signal and the robot’s actions, making imitation learning more effective. To implement all these requirements, we define several losses in the motion planning process to ensure the generation of high-quality demonstrations.


Finally, to distill these demonstrations into a visuomotor policy, we employ a 4D imitation learning approach that incorporates both human and object vision inputs, along with coordinated base-arm action outputs. Unlike previous works~\cite{Wang_2024_CVPR,christen2023learning} that focus solely on hand-object point clouds, our approach also includes the human body as input. To address scale differences between body and object point clouds, we apply set abstraction layers with varied sampling radii. The resulting features are merged into a global representation and decoded using an MLP to generate coordinated base-arm movements, which are essential for controlling a mobile robot.

We evaluate our learned policy in both simulators and the real world. Notably, without relying on human mocap assets or real-world robot demonstrations, our method outperforms baselines in all settings, achieving at least \textbf{+15\%} improvement in success rate.
Our experiments demonstrate that scaling up demonstration size and enriching scene variety significantly enhances policy generalizability. 
Furthermore, the automatically generated scalable demonstrations improve safety and accuracy, reducing collisions by about 1/3 and increasing success rates by 11.6\% while facilitating skill transfer to real mobile robot systems.

In summary, the key contribution of this paper is a novel, automated framework that scales learning for H2MR handover, consisting of three main components: i) a scalable pipeline for generating diverse synthetic full-body human motion data for the handover task, ii) an automatic pipeline that generates safe, imitation-friendly demonstrations for vision-based closed-loop control, and iii) a 4D imitation learning method for distilling large-scale demonstrations into closed-loop policies with base-arm coordination.


\section{Related Work}
\label{sec:related}

\subsection{Human-to-Robot Handovers}
Recent advancements in human-robot handovers \cite{DUAN2024100145,Wang_2024_CVPR, corsini2022nonlinear} have been fueled by the growing interest in human-robot interaction~\cite{obaigbena2024ai, MORIUCHI2024103682} and the availability of large datasets \cite{chao2021dexycb, ye2021h2o, fan2023arctic, liu2022hoi4d, deitke2023objaverse,deitke2024objaverse} capturing hand-object interactions. Grasping and dynamic motion planning~\cite{zhang2023flexible, yang2021reactive, fang2023anygrasp} are potential solutions but often face limitations in motion flexibility and large-scale datasets.
HandoverSim~\cite{chao2022handoversim} and GenH2R-Sim~\cite{Wang_2024_CVPR} provide physics-based environments and benchmarks for H2R handovers. By leveraging mocap data or large synthetic datasets, these platforms enable training learning-based policies~\cite{christen2023learning, christen2023synh2r, Wang_2024_CVPR}. However, these methods focus on fixed-base H2R handovers and do not consider large workspaces. \cite{10608385} attempts to decouple object pose estimation from planning, but it lacks scalability and does not model the environment comprehensively. In contrast, our proposed MobileH2R framework integrates human information, object data, and temporal context, solving the mobile handover task in an end-to-end manner. It incorporates scalable asset and demonstration generation processes, along with an efficient imitation learning approach.

\begin{figure*}[t]
  \centering
  \includegraphics[width=2\columnwidth]{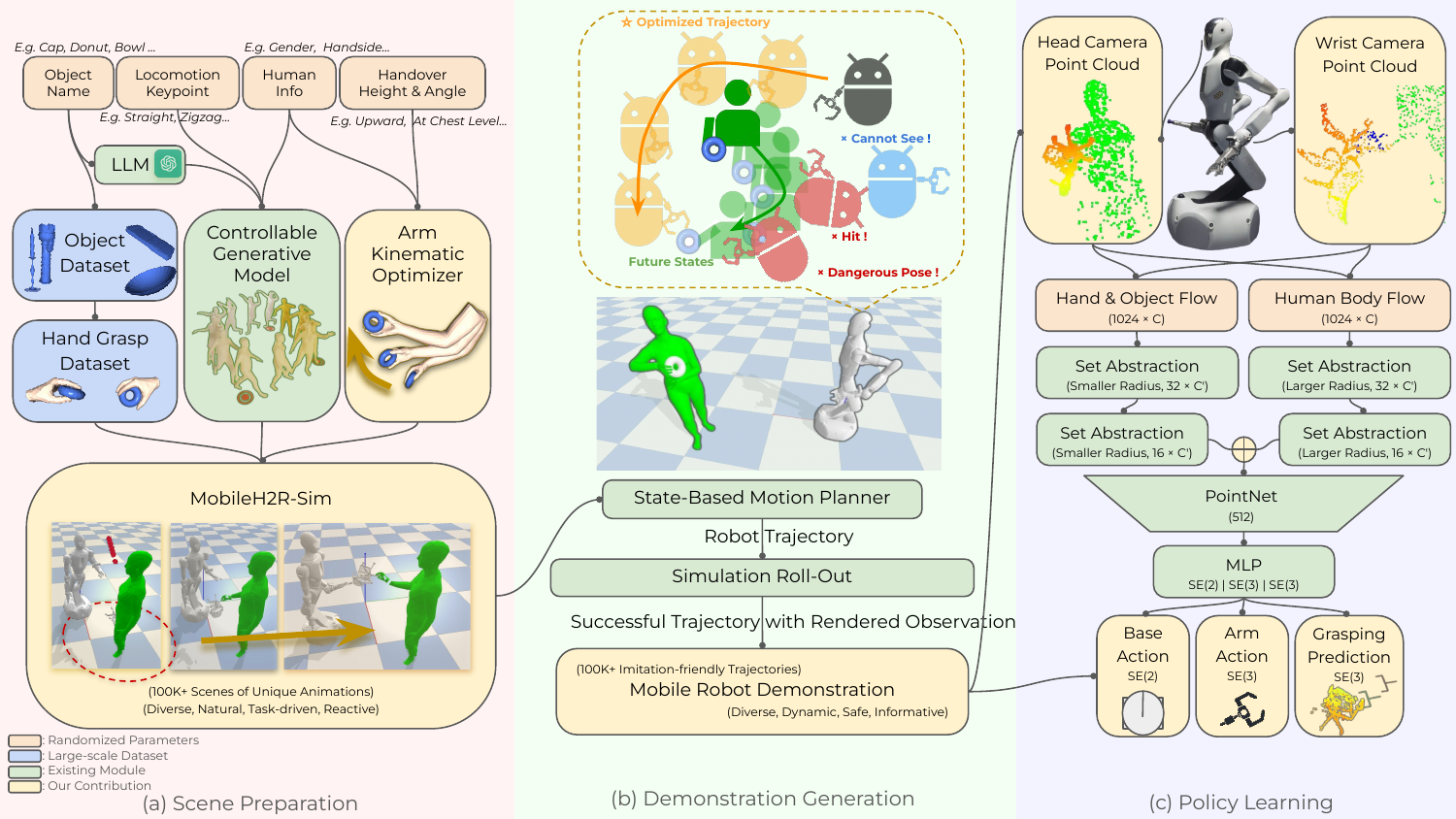}
  \vspace{-2pt}
  \caption{\textbf{The overview of our framework. } 
  First, we propose an automatic pipeline to scale up synthetic and diverse full-body motion data for the handover task by integrating various synthetic digital asset libraries, generative models, and useful toolkits. Second, we introduce an automatic pipeline to scale up mobile robot demonstrations for safety and imitation-friendliness. Our approach aims to avoid collisions while enhancing the vision-action correlation through carefully designed loss functions. Third, we employ a 4D imitation learning policy to learn 9D coordinated arm-base actions. We process point clouds of both objects and human bodies by modified PointNet++.}
\label{fig:overview}
\vspace{-0.4cm}
\end{figure*}   

\subsection{Mobile Robot Manipulation}
Many real-world tasks, like human-to-mobile robot handovers, require coordinated body movements while manipulating objects~\cite{liu2024visual,wang2024quadwbggeneralizablequadrupedalwholebody}. Many recent approaches separate navigation and manipulation using semantic controllers or motion planners~\cite{mullen2024towards,xiang2024language,xia2020relmogen}. ~\cite{yokoyama2023asc, li2020hrl4in, jauhri2022robot} address long-horizon mobile manipulation tasks by leveraging predefined navigation and manipulation skills, typically through iterative or two-stage methods.
However, these methods often overlook the need for coordinated arm and base actions. While some researchers~\cite{yang2023harmonic} address human-free tasks with reinforcement learning for mobile manipulation, our focus on human interaction requires ensuring safety. Therefore, we choose imitation learning with safe demonstrations to learn coordinated arm-based actions for human safety.

\subsection{Scaling up Demonstrations for Imitation}
Imitation learning from large-scale demonstrations has emerged as a popular approach for robot learning. These demonstrations can be sourced from various channels, including real-world collections~\cite{song2020grasping,o2023open}, foundation models~\cite{ha2023scaling,wang2023robogen,wang2023gensim}, non-robotics datasets~\cite{grauman2022ego4d}, and generating through Task and Motion Planning (TAMP)~\cite{garrett2021integrated, mcdonald2022guided, dalal2023imitating}. GenH2R~\cite{Wang_2024_CVPR} has adapted the TAMP framework for dynamic handover scenarios in human-to-robot (H2R) interactions. In this work, we focus on a more challenging task: H2MR handover, where both the human and robot behaviors are more dynamic and diverse, necessitating the generation of safe and imitation-friendly demonstrations.
\section{Method}
\label{sec:method}

For the generalizable human-to-mobile-robot handover task, we focus on a robot with a mobile base, controlling the movement of the base and the robotic arm.
The robot is equipped with a head camera and a wrist camera for visual input. When the robot is positioned farther from the human, the head camera is mainly used for perception. As the robot moves closer, the wrist camera supplements the input to overcome occlusion caused by the arm.
The model learns 6D control actions (3D translation and 3D rotation) for the robot gripper and 3D control actions (2D translation and 1D rotation) for the base, using segmented point cloud data of the human, hand, and object from both cameras. The framework is shown in Figure~\ref{fig:overview}.

\subsection{MobileH2R-Sim}
\label{sec:simulation}

There are two prime factors of large-scale synthetic human activities to resemble ground-truth distribution, i.e., diversity and fidelity (human-likeness). For task-specific human-robot interaction(HRI), like handover, two extra factors are worth taking into consideration, i.e., specialization and interactivity. The human agent in human activities synthesis requires full awareness of the specific HRI task and interacts with the robot agent in the simulated environment.

Conventional methods for collecting human trajectories are often unsuitable for sophisticated HRI tasks. Existing large-scale datasets, such as Human3.6M~\cite{h36m_pami}, AMASS~\cite{AMASS:ICCV:2019}, and MotionX~\cite{lin2024motionxlargescale3dexpressive}, lack task-specific focus. While motion capture or tracking techniques can be applied to specialized tasks, they are difficult to scale. The randomized trajectory generation approach, as implemented in GenH2R, can produce simple hand-object motions but cannot generate human-like motions. To address these challenges, we introduce MobileH2R-Sim, a simulation environment specializing in full-body motion synthesis for the H2MR handover task.
The generation process is divided into two phases. During the pre-handover phase, our approach generates diverse, task-independent full-body movements. During the handover phase, we generate arm-only movements to smoothly transition from the previous actions and initiate the object transfer. Additionally, we incorporate an interactive design that responds to the robot's actions, facilitating seamless transitions between the two phases, rather than relying on a static animation replay.

To generate diverse full-body movements during the pre-handover phase, we employ controllable generative models that distill a human motion prior from large-scale datasets. Specifically, we use Guided Motion Diffusion (GMD)~\cite{karunratanakul2023guidedmotiondiffusioncontrollable} trained on the AMASS dataset, generating full-body actions based on prompts and offers significant advantages over other methods. Since our goal is not to generate physically feasible motions for simulation, but rather to provide diverse human motion for H2MR handover scenarios, this approach avoids the scalability limitations of real-world data.  

For the arm-only movement during the handover phase, we adopt a task-specific local synthesis approach that introduces sufficient randomness. Unlike the full-body motion synthesis, this method focuses on the hand and arm movements involved in holding and transferring the object. The process consists of three stages: (1) identifying a feasible region in space for the handover and randomly sampling a point within it as the final handover position; (2) solving for the arm’s joint configuration at the final position using a kinematic optimizer that accounts for human joint constraints and movement habits; and (3) selecting an appropriate arm trajectory by calculating the difference between the initial and final positions, followed by interpolating intermediate frames to ensure smooth motion.

Another critical component is the interactive design, which allows for human responsiveness to the robot within the simulation to mimic the real world. When the robot is at a greater distance from the human, the human remains in the pre-handover phase, performing full-body movements without engaging in the handover. The phase transition is triggered when the robot approaches within a specific range or when the pre-handover time limit is reached. Once the handover movement is completed, all motion ceases, marking the end of the mobile handover task.

Several technical enhancements are incorporated into our approach. We utilize large digital asset libraries such as ShapeNet~\cite{chang2015shapenet}, which provides 8,836 synthetic objects, and Acronym~\cite{eppner2021acronym}, which offers corresponding grasping pose sets. To capture the nuanced hand poses necessary for a secure grip, we leverage DexGraspNet~\cite{wang2023dexgraspnetlargescaleroboticdexterous}. Furthermore, to ensure that the full-body movements generated by the generative model are object-aware, we resort to the large language model~\cite{openai2023gpt4} to generate motion prompts that align with the semantic attributes of the object.

\subsection{Safe and Imitation-friendly Demonstration}
\label{sub:demo}

\begin{figure}[t]
\centering
  \includegraphics[width=0.9\columnwidth]{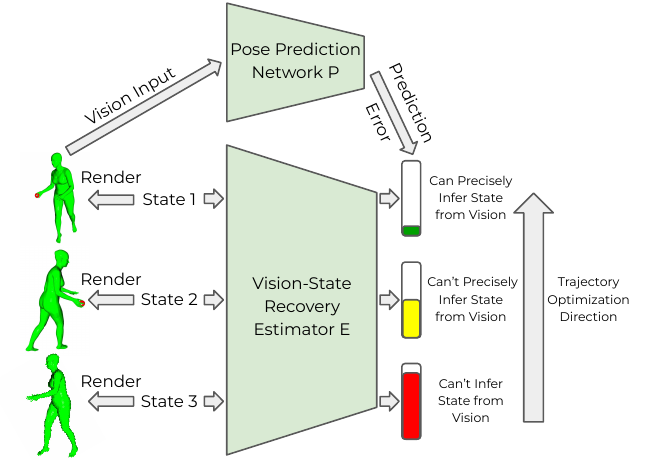}
  \caption{\textbf{Visualization for the vision neural loss.} The Pose Prediction Network takes vision inputs and predicts the object pose. The prediction error is defined as the vision neural loss. The Vision-State Recovery Estimator takes states as input and estimates the vision neural loss, guiding the state-based trajectory optimization towards imitation-friendly demonstration generation.}
\label{fig:two_strategies}
\vspace{-0.4cm}
\end{figure}

GenH2R~\cite{Wang_2024_CVPR} has shown that distillation-friendly demonstrations are crucial for vision-based closed-loop policy learning in fixed-base handover. 
Due to the robot's increased freedom and the resulting safety concerns, adapting this to mobile-base handover is challenging and requires finer-tuned strategies. Therefore, we aim to address two key questions in scalable demonstration generation for mobile-base handover: how to ensure safe human robot interaction and how to generate imitation-friendly demonstrations in mobile environments. 

Specifically, CHOMP~\cite{ratliff2009chomp}, an optimization-based planner is used as our expert planner, which is commonly applied directly or indirectly in recent handover research~\cite{Wang_2024_CVPR,christen2023synh2r,christen2023learning} to optimize robot trajectories. This planner offers simplicity and flexibility, allowing us to refine the initial trajectory through gradient descent to meet various objectives. 
To support this planner, we provide privileged scene knowledge, including the object’s 6D pose, hand and human poses, and candidate target object grasps generated via physics simulation~\cite{eppner2021acronym} at any moment.

For safety, we employ future obstacle avoidance to prevent collisions between the human and the robot over a time window, rather than just at the current timestamp. Traditional obstacle losses focus on immediate collisions, typically relying on the human's signed distance field and the robot's collision points at the current time, which may result in unavoidable collisions at future timestamps. In contrast, by utilizing oracle information, we are able to foresee the future positions of both the human and the robot, which enables us to calculate the obstacle loss at each future moment, proactively avoid potential collisions, and generate smoother and more natural robot movements.

In addition, we constrain the robot's final position, ensuring it stops in front of the person for a visible, face-to-face handover. This approach helps avoid movements that, while technically collision-free, could feel unsafe to the human—such as when the robot stands behind the person and extends its arm around to grasp an object from the front.

To ensure imitation-friendly learning for the visuomotor policy, we must ensure that the vision-to-action mapping is learnable. Since the demonstration generator essentially models the mapping between oracle states and actions without considering vision, our key insight is that if vision can effectively recover the state, the association between vision and action can be enhanced, making imitation learning easier. For example, if an object is occluded due to camera limitations, the demonstration may still succeed, but the policy cannot effe   ctively learn from this visual input for imitation.

To establish a clear vision-to-state-to-action pathway, we need to identify crucial state information that links the vision-action correlation. We find that object pose is a key oracle because knowing the object's position enables effective grasping actions. If we can recover the object's pose from vision, we can bridge the vision-to-state-to-action learning path.
Specifically, we train a pose prediction network \( P \) that takes vision input and predicts the object pose, defining the vision neural loss as the difference between the predicted pose and the ground truth pose. If the vision input accurately recovers the object pose, the loss is low.
Unfortunately, visual inputs rendered from simulations (such as PyBullet~\cite{coumans2021} or IsaacGym~\cite{makoviychuk2021isaac}) are often non-differentiable, meaning that the neural loss cannot propagate gradients for trajectory optimization. To address this, we trained a vision-state recovery estimator \( E \), which estimates the vision neural loss defined by \(P\) for each state. For state-vision pairs, \( E \) takes states as input, and predicts the vision neural loss for the vision inputs given by \( P \). Now vision neural loss estimated from \( E \) can propagate gradients to the states, allowing for trajectory optimization, as shown in Figure~\ref{fig:two_strategies}.

We can also add some heuristic designs to ensure visibility with the aim of enhancing the vision-action correlation, such as keeping the camera constantly oriented towards the object. However, based on experimental results, overly hacky designs often fail to capture the subtlety of the vision-action relationship, disrupt the robot’s normal posture and trajectory, and lack generalizability. In contrast, our neural vision loss, by modeling the state, implicitly optimizes a better vision-action correlation.

\subsection{Imitation  for Coordinated Base-Arm  Actions}
\begin{table*}[!ht]
\centering
\small
\begin{tabular}{c|ccc|ccc|ccc}
\hline
\multicolumn{1}{c|}{\multirow{2}{*}{}}                                           & \multicolumn{3}{c|}{m0}            & \multicolumn{3}{c|}{n0}                      & \multicolumn{3}{c}{s0}                      \\                                           &  Success          & Time   &  AS       &  Success         & Time   &  AS & Success          & Time  &  AS          \\ \hline
~~~ Grasp Selection + Trajectory Planning ~~~ & 40.20 &  11.75 &  8.70 & 34.80 & 12.83 & 5.02 &  40.97 &  9.61 & 14.71  \\

GenH2R & 4.80 &  6.93 &  2.58 & 3.10 & 8.66 & 1.31 & 40.97 & 4.93 & 27.5 \\

GenH2R (reprod.)  & 46.80 & \textbf{6.58}  & 26.27  & 32.90 & 7.03 & 17.48  & 61.11 & \textbf{4.67} & 42.09 \\

\hline



\quad Ours \quad & \textbf{63.80}  & 6.82 & \textbf{34.81}  &\textbf{ 53.40} &\textbf{ 6.94} & \textbf{28.68} & \textbf{77.78} & 5.23 & \textbf{50.65}\\ 

\hline

\hline
\end{tabular}
\vspace{-1mm}
 \caption{\textbf{Evaluation on different methods. } We compare our method against baselines across three test sets: the relatively simple human-involved scenario "m0", complex scenarios "n0", and real mocap data "s0". The "time" metric combines both computation and execution time. Since computation time varies depending on GPU and CPU configurations, we standardized it using an idle RTX 3090 with 32 CPU cores. Our policy outperforms the baselines in success rate and average success, while maintaining relatively low time cost.}
 \vspace{-1em}
 \label{tab:main_exp}
\end{table*}
Recent human-to-robot handover approaches focus mainly on the interaction between the human hand and the robot's end-effector, often neglecting contextual factors such as the human's motion and posture. Many methods crop only the hand and object point clouds~\cite{yang2021reactive,christen2023learning}, which could introduce safety risks. Additionally, state-of-the-art models~\cite{Wang_2024_CVPR} typically focus on 6D egocentric actions at the arm’s end-effector, ignoring the robot's base coordination—critical for mobile robots. To address these challenges, we propose a scene-fused 4D imitation learning approach that integrates full-body coordination and vision-action dynamics for more holistic human robot interactions.

To fully leverage the scene information, we combine the point clouds from the front and wrist cameras, which are segmented into objects, hands, and human body components, and align them with the robot’s end-effector frame. Our policy uses these point clouds as inputs and outputs an egocentric 6D action for the robotic arm and an egocentric 3D action for the  mobile base. Inspired by GenH2R~\cite{Wang_2024_CVPR}, we balance the policy’s 4D learning capabilities with real-world inference speed by reconstructing the point cloud at each timestamp and calculating flow information between consecutive point clouds using the Iterative Closest Point registration algorithm. This method approximates the previous timestamp’s point cloud coordinates, enhancing the current point cloud features with 4D information. To extract global features from point clouds enriched with 4D information, we typically use a PointNet++~\cite{qi2017pointnet++} encoder. However, due to the differing spatial scales between the human body and hand-object point clouds, using a single set abstraction layer for the entire point cloud often causes the FPS algorithm to prioritize points from the human body. To address this, we apply set abstraction layers with distinct sampling radii for the human body and hand-object regions. In the final PointNet layer, we concatenate these features to obtain a unified global feature representation.

To ensure coordinated movement between the base and the robotic arm, we directly input the unified global feature into an MLP layer, which decodes base and arm movements simultaneously. The output is supervised using coordinated movements from the demonstrations, resulting in the losses \( \mathcal{L}_{\text{base}} \) and \( \mathcal{L}_{\text{arm}} \). Additionally, an auxiliary task is employed to predict the object's grasping pose, enabling the robotic arm to better anticipate the target position. Our final loss function is defined as \( \mathcal{L} = \lambda_1 \mathcal{L}_{\text{base}} + \lambda_2 \mathcal{L}_{\text{arm}} + \lambda_3 \mathcal{L}_{\text{pred}} \), where \(\lambda_1\), \(\lambda_2\), and \(\lambda_3\) are used to balance the losses. This distillation approach enables our policy to naturally execute coordinated base-arm actions, generalizing effectively to dynamic scenarios.

\section{Experiments}
\label{sec:experiments}

\textbf{Dataset.} 
(1) As described in Section~\ref{sec:simulation}, Our data generation pipeline enables the creation of a large number of complex synthetic H2R handover scenes, incorporating custom human animations and 8,836 diverse objects from ShapeNet~\cite{chang2015shapenet}. We designed two setups, ``m0'' and ``n0'', to represent different levels of diversity in human motion generation. In ``m0'', the human approaches straightforwardly to hand over the object, simulating a relatively simple scenario. In ``n0'', the human performs the handover with more complex body motions, trajectories, and speeds, such as sitting, running, descending stairs, or dancing, to mimic more varied real-world handover situations. Each setup includes 100k training scenes and 1k testing scenes.
Each handover action includes a 6s pre-handover phase with natural movements, followed by a 1.05s handover phase for object transfer. The transition to the handover phase occurs when the human-robot distance falls below a 1m threshold or when the pre-handover phase has elapsed.
(2) To incorporate more real-world handover scenes into MobileH2R-Sim for evaluation, we use the DexYCB~\cite{chao2021dexycb} dataset from HandoverSim~\cite{chao2022handoversim}, which provides 1,000 real-world H2R handover scenes and 20 objects. As this dataset captures only hand and object motion data, we add a human body model to align with our setup without altering the existing hand or object trajectories. We use the simultaneous ``s0'' setup, containing 720 training scenes and 144 testing scenes.

\noindent\textbf{Metrics.} We adhere to the GenH2R~\cite{Wang_2024_CVPR} evaluation protocol. A successful handover requires securely grasping the object from the human hand without any collision. Failure cases include human contact, object drop, and timeout ($T_{max}=15s$). We report the success rate and the time. 
It is noted that the ``time'' metric consists of inference time (policy action computation) and execution time (action execution in simulation). While GenH2R only considers inference time, we include both, offering a more accurate reflection of real-world performance.
To evaluate both success rate and completion efficiency, we use AS (Average Success), akin to AP (Average Precision):
\begin{equation}
\text{AS} = \int_0^1 \text{Success}(t) \, \mathrm{d}t
\label{equ:ap}
\end{equation}
where $\text{Success}(t)$ is success rate considering only successful cases within $t \cdot T_{\text{max}}$. This method can better evaluate success-time relations which is suitable in HRI scenarios.


\subsection{Evaluation on Different Methods}
\label{sub:evaluating_on_different_method}

\textbf{Setup.} To ensure a fair comparison between baseline methods and our imitation policies powered by imitation-friendly demonstrations, we train all methods on the 10k ``n0'' training scenes and evaluate them on three testing sets: ``m0'', ``n0'', and ``s0''. 

\noindent\textbf{Baselines.} ``Grasp Selection + Trajectory Planning'' is a non-end-to-end, straightforward method. The point cloud input is used to predict the target grasping pose, either via GraspNet~\cite{fang2020graspnet,fang2023robust} or by matching current visual input with the object's ground truth grasping pose~\cite{eppner2021acronym}. Based on it, motion planning is performed at each time step. Although time-consuming, this baseline is widely adopted for its ease of implementation.
GenH2R is an end-to-end approach for fixed-base handover tasks, converting the current hand-object point cloud into robotic arm motions. However, it lacks a full-body robot and only outputs 6D arm actions. We can extend it to base-arm actions by solving full-body inverse kinematics. Additionally, GenH2R's environment does not include humanoid interaction. We either use the pre-trained GenH2R model or retrain it in ``n0'' (GenH2R(reprod.)) to learn 6D arm actions with the mobile base and incorporate collision-free demonstrations. 

\noindent \textbf{An end-to-end framework models the HRI scenario more effectively.} Compared to ``Grasp Selection + Trajectory Planning'', our method achieves a 26.3\% higher success rate, reduces time by 5.06s, and increases average success by 28.6\% on average. The non-end-to-end method fails mainly due to collisions with human, as it plans based only on the current step and doesn't anticipate human movements to avoid them in advance. Frequent planning steps also increase computation time, with our policy requiring about 0.003s for inference, while motion planning usually takes over 0.1s. These results show that an end-to-end framework better captures the global context, though more complex methods like MPC could consider more factors, they often lack scalability.

\begin{figure*}[t]
    \includegraphics[width=0.9\linewidth]{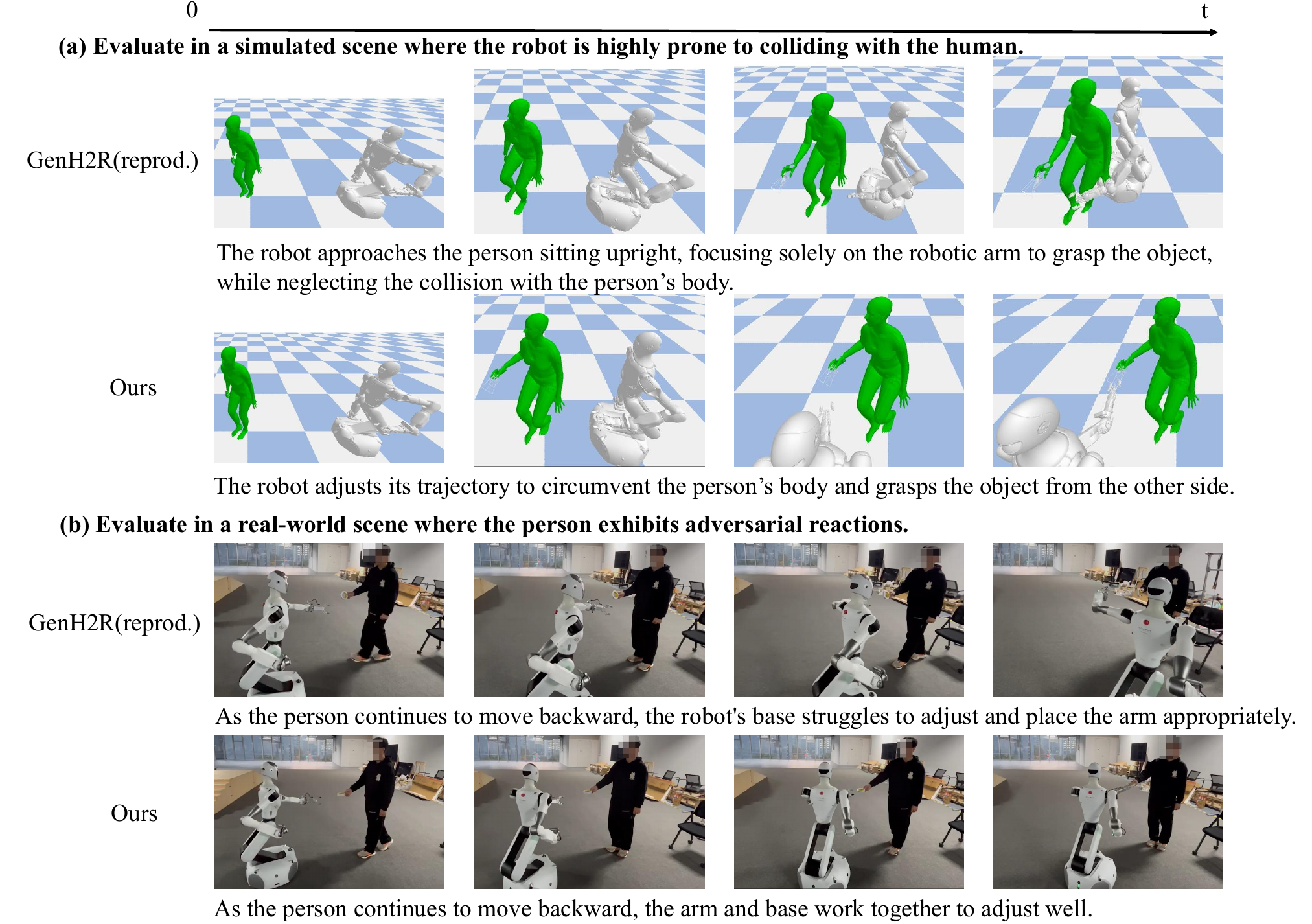}
    \centering
    \vspace{-0.5em}
    \caption{\textbf{Qualitative results}. We compare different methods in detail in the simulated scene and the real-world scene.}
    \label{fig:real_world}
    \vspace{-1em}
\end{figure*}

\noindent\textbf{Human involvement and mobility are crucial.} For the baseline GenH2R, handover success is challenging, with frequent human collisions in the ``m0'' and ``n0'' test sets due to the lack of awareness of complex human motion and intent. In contrast, GenH2R(reprod.) benefits from demonstrations that implicitly avoid collisions, improving success rates by 42.0\%, 29.8\%, and 20.1\% across the three test sets. However, it only learns 6D arm actions and computes 3D base actions via inverse kinematics.
Our approach, which incorporates human information and considers base-arm coordination, achieves further success rate improvements of 17.0\%, 20.5\%, and 16.7\%. This shows that learning base-arm coordination is essential. A strategy suitable for a fixed-base arm cannot be directly applied to a mobile robot without accounting for its specific configuration. We also show visualizations in Figure \ref{fig:real_world} (a).

\subsection{Evaluation on Data Scaling}
\label{sub:evaluating_Scaling}


\definecolor{darkbrown}{RGB}{120, 60, 30}
\definecolor{deepolivegreen}{RGB}{85, 107, 47}
\definecolor{green}{RGB}{0, 255, 0}




\begin{table}[t]
\centering
\setlength{\tabcolsep}{3mm}
\small
\begin{tabular}{cccc}
\hline
\textbf{Num. of Demos / training set}              & m0 & n0 &  s0\\ \hline
100k, n0            &  65.6$\uparrow$     &   56.6$\uparrow$     &  82.6$\uparrow$  \\
1k, n0          & 53.4$\downarrow$ & 42.3$\downarrow$ & 57.6$\downarrow$ \\
10k, n0   & 63.8 & 53.4  & 77.8\\
10k, m0          &  62.0$\downarrow$ & 45.3$\downarrow$ & 76.4$\downarrow$ \\
10k, s0          & 35.3$\downarrow$  & 22.0$\downarrow$&  34.0$\downarrow$      \\
\hline
\end{tabular}
\vspace{-1mm}

\begin{tikzpicture}[overlay, remember picture]
    \draw[decorate, decoration={brace, amplitude=6pt}, color={rgb,255:red,120; green,60; blue,30}, thick] 
        ([yshift=-6pt] -2.55, 1.34) -- ([yshift=6pt] -2.55, 1.99) 
        node[midway, xshift=-0.5cm, red] {};
    \draw[decorate, decoration={brace, amplitude=6pt}, color={rgb,255:red,85; green,107; blue,47}, thick] 
        ([yshift=-6pt] -2.85, 0.54) -- ([yshift=6pt] -2.85, 1.19) 
        node[midway, xshift=-0.5cm, blue] {};
\end{tikzpicture}
\vspace{-11pt}
\caption{\textbf{Evaluation on varying demonstration numbers and Assets.} We compare the policy's success rate across three testing sets, examining the impact of different demonstration sizes within \textcolor{darkbrown}{brown braces} and asset variations within \textcolor{deepolivegreen}{green braces}.}
 \vspace{-1em}
 \label{tab:scaling}
\end{table}

As shown in Table~\ref{tab:scaling}, we evaluate the impact of scaling from two perspectives: demonstration scaling and asset scaling.

When comparing different demonstration sizes trained on the same dataset ``n0'', We observe that using 100k demonstrations increases the success rate by 3.3\% on average compared to using 10k demonstrations, while using 1k demonstrations leads to a 13.9\% decrease in success rate.
These results highlight the importance of dataset size in our imitation learning paradigm. Our large-scale synthetic data and efficient demonstration pipeline mitigate concerns about limited datasets affecting generalization.

On the other hand, using the unscalable mocap dataset ``s0'' results in a significant drop in the success rate by 34.6\% on average.
When trained on the simpler human-involved ``m0'' dataset, the policy performs well on the simple testing set, with a small drop in success rates of 1.8\% and 1.4\% for ``m0'' and ``s0''. However, in more complex scenarios, the success rate decreases by 8.1\%. This shows that the complexity and diversity of human assets are crucial for the policy’s ability to handle more challenging tasks. 

The above results validate our assumption that, for HRI tasks, \textbf{increasing the number of demonstrations and the diversity of assets is key to improving policy performance.} This also highlights the value of our scalable framework.

\subsection{Evaluation on Demonstration Strategies}

\begin{table}[t]
\centering
\setlength{\tabcolsep}{3mm}
\small
\begin{tabular}{cccc}
\hline
\textbf{Demo strategies}              & m0 & n0 &  s0\\ \hline
w/o future obstacle avoidance           & 58.7   & 48.5 &  75.7  \\
w/o final pose constraints & 58.4  & 43.2 & 69.4 \\
w/o imitation-friendly loss          & 56.8  & 46.4 & 56.9 \\
with camera orient-based loss    & 54.0 & 42.8 & 56.0 \\
safety concern + vision neural loss   &\textbf{ 63.8} & \textbf{53.4} & \textbf{77.8}\\
\hline
\end{tabular}
\vspace{-1mm}
\caption{\textbf{Evaluation on different demonstration generation strategies.} We compare the policy's success rate across three testing sets. They all trained on ``n0''.}
 \vspace{-1em}
 \label{tab:demo_method}
\end{table}

As shown in Table~\ref{tab:demo_method}, we evaluate different demonstration strategies in terms of safety and imitation-friendliness. 

Without future obstacle avoidance, the policy’s success rate decreases by an average of 4.0\%, and the human contact rate increases from 10.9\% to 16.1\%. Without final pose constraints, the success rate decreases by an average of 8.0\%, and the contact rate rises to 19.8\%. These results highlight that demonstrations with a simple safety design can significantly reduce human collisions, which are the most critical failures to avoid, as compared to failures such as object drops or timeouts.

For the imitation-friendly purpose, we find that without encouraging easy vision-to-action learning, the success rate drops by 11.6\%. The loss for training the policy is much higher when dealing with irrational vision-action pairs. When we manually design an orientation loss for the robot camera to keep it continuously focused on the object, the policy performs worse than one without the orientation constraint. We check the generated demonstrations and notice that many of them attempt to align the object’s angle, leading to stagnation and discontinuity in actions, which results in even more irrational vision-action pairs. Therefore, generating distillation-friendly demonstrations requires a balance between various considerations.


\subsection{Ablations}
As depicted in Table~\ref{tab:policy_ablation}, we make ablations on different designs for our imitation policy.
The absence of flow information leads to a 12.1\% decrease in success rate, emphasizing the importance of utilizing 4D data.
Without human information, the success rate drops by 12.5\%, highlighting the need to consider human-related context in HRI tasks.  
When decoding the arm and base actions separately using two independent networks (e.g., two PointNet++), we observe a dramatic 17.8\% decrease in success rate, demonstrating the necessity of decoding both actions simultaneously.

\label{sub:ablations_on_policy}

\begin{table}[t]
\centering
\setlength{\tabcolsep}{3mm}
\small
\begin{tabular}{cccc}
\hline
\textbf{Methods}              & m0 & n0 &  s0\\ \hline
w/o flow             &   ~~58.0~~ &   ~~47.2~~  & ~~53.5~~ \\
w/o human            &  51.3  &   48.7     & 57.6 \\
w/o coordinated action      & 51.0    & 41.7 & 48.9 \\
Ours      & \textbf{63.8} & \textbf{53.4} & \textbf{77.8}  \\
\hline
\end{tabular}
\vspace{-1mm}
\caption{\textbf{Ablations on different policy designs.} We conduct ablations on various modules, including flow information, human information, and coordinated base-arm actions.
}
 \vspace{-1em}
 \label{tab:policy_ablation}
\end{table}

\subsection{Real World Experiments}

\begin{table}[t]
\centering
\small
\begin{tabularx}{1.0\linewidth}{ccc}
\hline
\textbf{Methods}              & ~~~\textbf{m0}~~~  & ~~\textbf{n0}~~~  \\ \hline
GenH2R(reprod.)~\cite{Wang_2024_CVPR} & ~~12 / 30 (40.0\%)~~ & ~~9 / 30 (30.0\%)~~ \\
Ours & 24 / 30 (\textbf{80.0\%}) & 19 / 30 (\textbf{63.3\%}) \\ \hline

\end{tabularx}
\vspace{-1mm}
 \caption{\textbf{Sim-to-real experiments.} We report the success rate of our method and GenH2R(reprod.) in two different settings.
}
 \vspace{-0.6cm}
 \label{tab:real_world}
\end{table}

\textbf{Sim-to-Real Transfer.} 
we deploy the models trained in the simulation on a real mobile robot, Galbot G1, equipped with an omnidirectional wheel chassis, a 7-DOF robotic arm, a head depth camera, and a wrist-mounted depth camera. We use SAM2~\cite{ravi2024sam2} to segment the point cloud data from the cameras, which is then fed into the policy to generate 9D egocentric actions for both the base and the left arm. The robot’s movements are controlled using a position controller. A user study compares our method against the GenH2R(reprod.) approach. Further details are provided in the supplementary material.

\noindent \textbf{User Study.} We compare our method with GenH2R(reprod.) across 6 objects in 2 different settings. In the simple setting, users directly hand over the object to mimic ``m0''. In the complex setting, users may sit, go downstairs, or perform adversarial actions to mimic ``n0''. The results are reported in Table \ref{tab:real_world} and visualizations are shown in Figure \ref{fig:real_world}(b). We observe that our model outperforms in completing the handover process across different objects and scenarios.


\label{sub_sim_to_real}
\section{Conclusion}
In this work, we present the MobileH2R framework for scaling the learning of generalizable human-to-mobile-robot (H2MR) handover. We introduce a new simulation environment for this task and a scalable pipeline for generating diverse synthetic full-body human motion data. Additionally, we propose an automatic method for creating safe, imitation-friendly demonstrations, enabling a 4D imitation learning approach to train coordinated base-arm actions. Our experiments show that generalizable handover skills can be developed in simulators using high-quality synthetic data, without the need for real-world data. We validate the approach both in the simulator and in the real world.
{
    \small
    \bibliographystyle{ieeenat_fullname}
    \bibliography{main}
}

\clearpage
\appendix

\twocolumn[\begin{center}
    \large\bfseries MobileH2R: Learning Generalizable Human to Mobile Robot Handover Exclusively from Scalable and Diverse Synthetic Data (Supplementary Material)\par
    \end{center}]
\addcontentsline{toc}{section}{Appendix}


The supplementary material provides additional details about various aspects of our method and experiments. Refer to the table of contents below for an overview. 
Section~\ref{supp_sec:more_method} offers further details on our framework and methodologies. Sections~\ref{supp_section:simulation} and~\ref{supp_sec:real_world} elaborate on the experimental setups and results in simulation and real-world scenarios, respectively. Section~\ref{supp_section:limit} discusses the limitations of our work and explores potential future research directions for human-to-mobile-robot handovers and broader human-robot interactions.

Additionally, we have attached \textbf{a video} that provides an overview of our method, along with extensive demonstrations in both simulation and real-world scenarios.

We are committed to contributing to the research community and will release our code publicly in the near future.

\etocdepthtag.toc{mtappendix}
\etocsettagdepth{mtchapter}{none}
\etocsettagdepth{mtappendix}{subsection}
{
  \hypersetup{
    linkcolor = black
  }
  \tableofcontents
}

\section{More Method Details}
\label{supp_sec:more_method}

\subsection{MobileH2R-Sim}

\begin{figure}[t]
\centering
  \includegraphics[width=1.0\columnwidth]{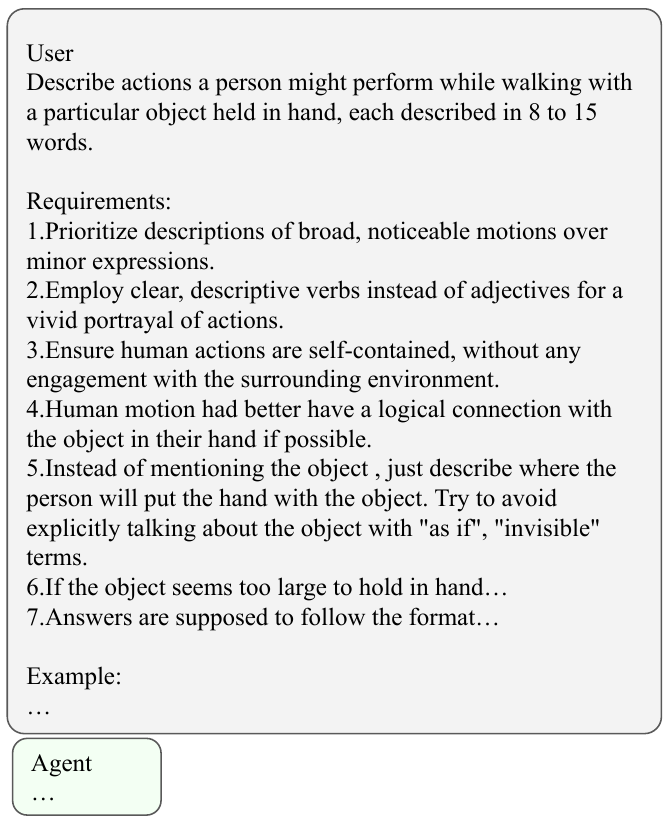}
  \caption{\textbf{Template prompt to LLMs} to generate direct object-aware motion description for controllable motion generator.}
\label{fig:prompt}
\end{figure}
\begin{figure}[t]
\centering
  \includegraphics[width=0.9\columnwidth]{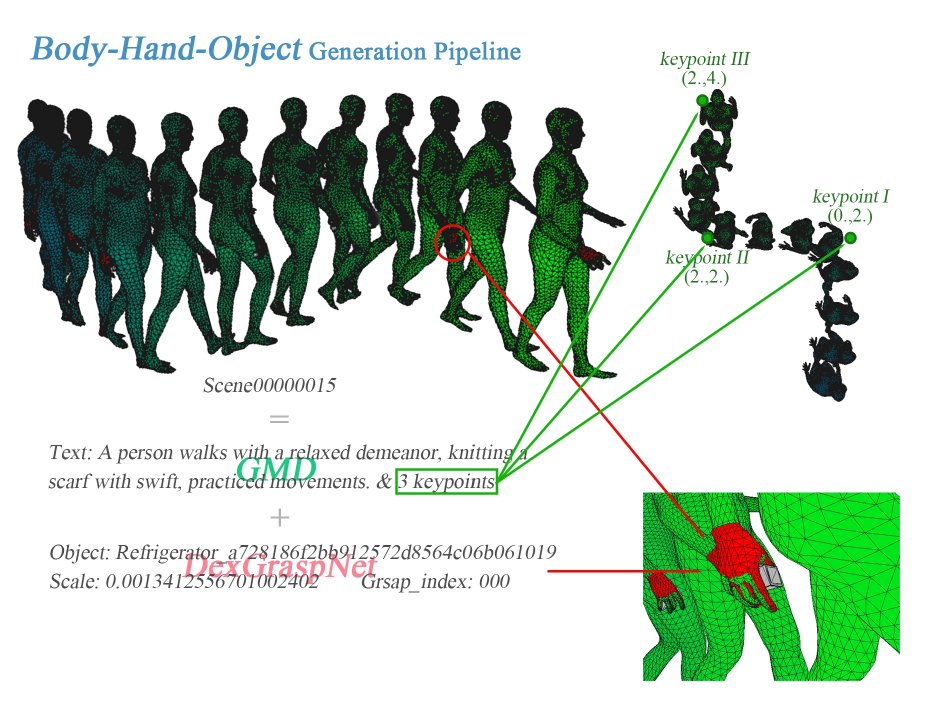}
  \caption{\textbf{Combine the body motion and the hand-object pose.} In our pipeline, the body motions are SMPL~\cite{SMPL:2015} parameters obtained from GMD~\cite{karunratanakul2023guidedmotiondiffusioncontrollable}, and the hand-object poses are MANO~\cite{MANO:SIGGRAPHASIA:2017} parameters obtained from DexGraspNet~\cite{wang2023dexgraspnetlargescaleroboticdexterous}. Here we combine these parameters with SMPL-X~\cite{SMPL-X:2019} model, as well as update the object poses relative to the human body.}
\label{fig:gmdex}
\end{figure}
\begin{figure}[t]
\centering
  \includegraphics[width=0.9\columnwidth]{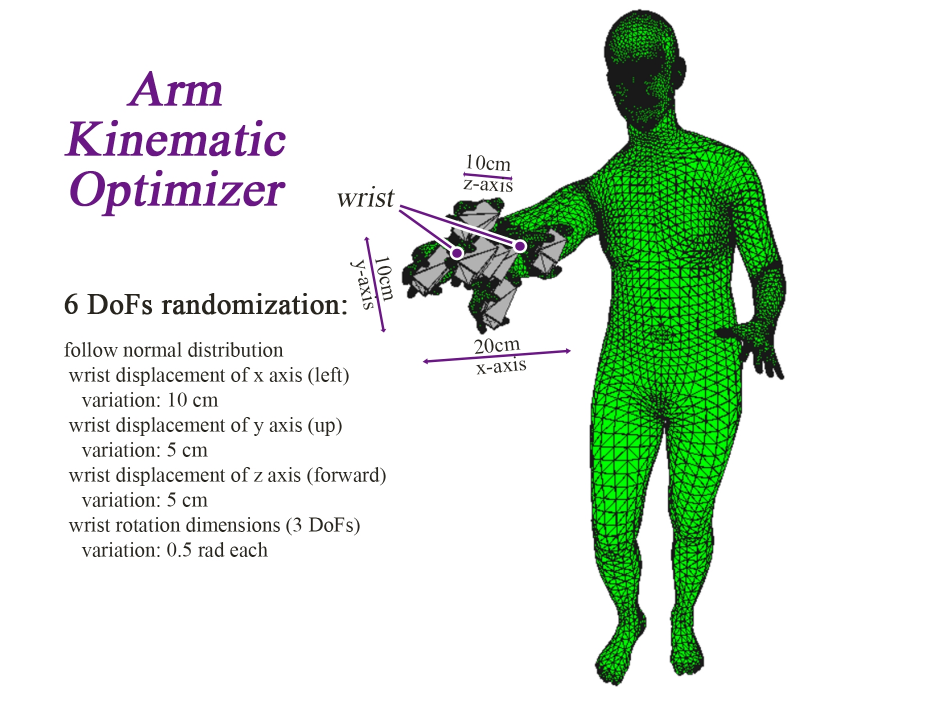}
  \caption{\textbf{Visualization for the arm kinematic optimizer.} Given the final pose of the hand, the optimizer determines the optimal hand joint parameters by optimizing under joint constraints and handover task priors. Practically, we add proper randomization to generate diverse data.}
\label{fig:handoverIK}
\end{figure}

The first innovation we would like to introduce in MobileH2R-Sim is its object-aware motion prompt generation. As our human-motion generator, GMD~\cite{karunratanakul2023guidedmotiondiffusioncontrollable}, claims the capability to generate semantically correct human motion trajectories, it is theoretically possible to compose diverse motion patterns according to different objects involved in the handover task. For instance, if it is a phone, we may imagine that one will hold the phone up to one's ear with one's hand before handing it over. Or if it is a bottle, a more natural action would be moving it to one's lip and taking a sip.

However, restrained by the problem setting of general motion synthesis and mainstream data annotation, motion diffusion models at the current stage including GMD, only accept direct description as guidance, lacking the capability to understand complex semantic implications. Besides, the information of the object cannot be directly obtained by the generative model as well. So we propose to involve a large language model, like GPT4~\cite{openai2023gpt4}, in the loop. With proper prompt~\ref{fig:prompt} design, we utilize the LLM to perform (1) conceptualizing actions of a person regarding a specified item and (2) describing specific postures of a person while performing the actions. With the output as the prompt to GMD, there are two significant benefits. Firstly, the generated motion is more diverse hence better encompasses the distribution of real human motion. Secondly, it becomes more likely to generate object-aware motion, making the scene more harmonious and realistic.

Another noteworthy aspect of MobileH2R-Sim is the strategy we adopted to integrate human motion with hand-object poses as shown in figure~\ref{fig:gmdex}. Building on the already obtained SMPL body model parameters from the motion process, we further utilize the SMPL-X model, which supports both body and hand modeling. We integrate hand poses derived from DexGraspNet, based on the MANO model, into the SMPL-X framework. Additionally, by calculating the mapping relationship between the hand joints of the MANO and SMPL-X models, we obtain a transformation matrix that aligns the hand with the body. This transformation allows the object pose to be projected into the relative pose within the body coordinate system.

For the arm motion during the handover phase, we implemented an optimizer based on joint constraints and task priors as shown in figure~\ref{fig:handoverIK}. Joint constraints include restrictions such as the elbow not bending backward and preventing body or arm mesh interpenetration. Task priors primarily ensure that the distribution of the final handover target points aligns with the task expectations. After obtaining the joint parameters for the arm's final frame through the optimizer, we generate the arm's lifting motion by applying appropriate interpolation. Similarly, we add the hand-object pair into this phase following the same scheme as discussed above.

After separately generating full-body motion and arm motion, we designed an effective interacting mechanism for switching between these two phases within the simulation environment. Initially, the human in the simulation only performs full-body motion. The transition from full-body to arm motion is triggered under one of the following conditions:(1)Proximity to the robot: When the distance between the robot and the human falls below a set threshold (0.5m), it is assumed that the human notices the robot's request to hand over an object. After a reaction time (approximately 200ms on average), the human begins executing the arm motion to hand over the object. (2) Completion of full-body motion: If the full-body motion has been completed (lasting a total of 6 seconds), the human will directly begin executing the arm motion to hand over the object. When both the full-body and arm motions are completed, or when the robot touches the object, the human stops moving and waits for the task to terminate. To address the uncertainty of when the arm motion will be triggered and to avoid the computational burden of real-time calculations, we precompute and cache arm motion at small intervals for each segment of full-body motion. When the arm motion is triggered, the simulation directly uses the cached arm motion for the subsequent time frame.

\subsection{Demonstration Workflow}

In our manuscript, we primarily focus on the design of loss functions for demonstration generation to ensure safety and imitation-friendliness. Here, we detail the overall demonstration workflow, with particular attention to overlooked components such as the forecasting window, grasp selection, and stopping distance.

\subsubsection{Step 1: Initialization}
Before starting the demonstration, we randomize the scene by applying a global translation and rotation to the human, hand, and object. This randomization serves as a form of data augmentation to enhance model generalization.

\subsubsection{Step 2: Plan Evaluation}
We set a predefined \textbf{replan interval}. Replanning is triggered under three conditions:
\begin{enumerate}
    \item The trajectory buffer is empty.
    \item The replan interval elapses.
    \item Reactive behavior occurs (e.g., transitioning from the pre-handover phase to the handover phase).
\end{enumerate}

If replanning is required, proceed to Step 3; otherwise, execute the current trajectory in Step 4. In our experiments, we set the replan interval to 1.5 seconds. This interval represents a tradeoff:
\begin{itemize}
    \item \textbf{Frequent replanning} slows down demonstration generation due to the computational overhead of replanning (Step 3) and may lead to trajectory discontinuities.
    \item \textbf{Infrequent replanning} risks mimicking GenH2R's~\cite{Wang_2024_CVPR} destination planning, which significantly reduces success rates.
\end{itemize}

\subsubsection{Step 3: Replanning}
During each replanning step, the planner uses oracle information from a defined foresee time window to predict the future states of the human, hand, and object. The planner then computes a new trajectory.
\textbf{Foresee time} plays a crucial role in planner performance, as it determines the planner's predictive capability. A shorter foresee time resembles dense planning, while a longer one approximates destination planning. Based on empirical testing, we fixed the foresee time to 1.5 seconds, ensuring consistent results across experiments.
Replanning involves two substeps:
\begin{enumerate}[label=\textbf{Step 3.\arabic*}]
    \item Generating the full-body robot grasping pose.
    \item Optimizing the trajectory.
\end{enumerate}

\textbf{{Step 3.1: Full-Body Robot Grasping Pose Generation}}
A naive approach would randomly select a grasping pose for the object, solve for an IK solution using the robot's URDF, and obtain a final grasping pose. However, this approach often results in collisions with the human or occlusions in the robot's view.

Instead, we preprocess grasping poses offline (provided by \textit{Acronym}~\cite{eppner2021acronym}) and rank them in descending order of distance from the human hand. The assumption is that a grasping pose farther from the human hand will generate a safer full-body pose when the robot is facing the human. Experimental results confirm this assumption.

We further constrain the robot's base position to a \textbf{sector-shaped area} directly facing the human, approximately 0.5–0.8 meters away. Instead of solving these constraints directly, we sample base positions and orientations within the sector. Once the base's 3 degrees of freedom (DoF) are determined, we attempt to solve IK using the ranked grasping poses. If an IK solution is found, it becomes the final full-body robot pose.

To accelerate this process, we prefilter grasping poses that collide with the human hand. This two-loop approach (outer loop for base sampling, inner loop for grasping pose selection) efficiently finds feasible solutions, as most valid poses are already facing the human.

\textbf{{Step 3.2: Trajectory Optimization}}
Given the initial and final grasping poses, we optimize the trajectory using the manuscript-defined obstacle loss and neural loss to ensure safety and distillation-friendliness.
\begin{itemize}
    \item The \textbf{neural loss} incorporates outputs from both the pose prediction network \(P\) and the vision-state recovery estimator \(E\), both implemented as simple 2-layer MLPs to maintain computational efficiency.
    \item Training \(E\) requires inputs in both the \textbf{end-effector frame} and \textbf{robot base frame}, including the human skeleton's keypoints, human pose, hand pose, and object pose. Since the trajectory is differentiable, gradients propagate through \(E\) to facilitate trajectory optimization.
\end{itemize}

\subsubsection{Step 4: Trajectory Execution}
The robot executes the optimized trajectory. If the trajectory is empty (e.g., due to IK failure), the robot remains stationary to ensure safety.

\textbf{Experimental Environment and Data Generation}
In our experiments, we utilized the PyBullet environment with Ray for multi-threading. Using an 8-GPU setup (Nvidia GeForce RTX 3090), we achieved a trajectory generation speed of approximately 1 second per trajectory. This setup ensures a success rate of over\textbf{ \textbf{80\%}}, allowing us to efficiently generate a large dataset.

We selected the majority of successful trajectories to serve as training data for subsequent imitation tasks. This approach enabled us to balance speed and accuracy while maintaining high-quality demonstrations.

\subsection{Imitation for coordinated Based-Arm Actions}

Our imitation network employs three loss functions: \( \mathcal{L}_{\text{base}} \), \( \mathcal{L}_{\text{arm}} \), and \( \mathcal{L}_{\text{pred}} \).

- Base Loss (\( \mathcal{L}_{\text{base}} \)):  
  This loss supervises the robot base's three degrees of freedom using the L1 loss function.

- Arm Loss (\( \mathcal{L}_{\text{arm}} \)):  
  Following the approach defined in~\cite{li2018deepim}, we use the predicted 6D pose to compute the coordinates of keypoints. The loss is then defined as the difference between the predicted and ground truth keypoint coordinates.

- Prediction Loss (\( \mathcal{L}_{\text{pred}} \)):  
  Similarly, \( \mathcal{L}_{\text{pred}} \) is defined using the difference between predicted and ground truth keypoint coordinates, following the same procedure as \( \mathcal{L}_{\text{arm}} \).

At each time step, we retrieve the point clouds from several previous time steps and apply the Iterative Closest Point (ICP) registration algorithm~\cite{rusinkiewicz2001efficient} to estimate the transformation matrices between the current point cloud and past point clouds in the world frame. 

While these transformations may introduce slight inaccuracies due to the incomplete nature of the point cloud inputs, they still provide sufficient flow information for each point. The estimated flow is then transformed back to the current egocentric frame, forming a crucial component of our feature representation.

In our experiments, we set the weighting coefficients as \(\lambda_1 = \lambda_2 = 1\) and \(\lambda_3 = 0.5\) and set the flow number as 3.
\section{Simulation Experiments Details}
\label{supp_section:simulation}

\begin{table*}[!ht]
\centering
\small
\begin{tabular}{c|cc|cc|cc}
\hline
\multicolumn{1}{c|}{\multirow{2}{*}{}}                                           & \multicolumn{2}{c|}{m0}            & \multicolumn{2}{c|}{n0}                      & \multicolumn{2}{c}{s0}                      \\                                           &  Success          & Contact / Drop / Timeout  &  Success          & Contact / Drop / Timeout  &  Success          & Contact / Drop / Timeout         \\ \hline



trained on s0 & 35.3  & 24.0 / 17.4 / 23.3 & 22.0 & 51.5 / 9.9 / 16.6 & 34.0 & 25.0 / 11.1 / 29.9 \\
trained on m0 & 62.0 & 7.3 / 18.8 / 11.9 & 45.3 & 26.8 / 16.3 / 11.6 & 76.4 & 9.0 / 9.7 / 4.9 \\
trained on n0 & 63.8 & 6.6 / 13.5 / 16.1 & 53.4 & 20.5 / 14.1 / 12.0 & 77.8 & 5.6 / 10.4 / 6.2 \\
\hline




\hline

\hline
\end{tabular}
\caption{\textbf{Evaluation on different training scenes.} 
We train our method on three training sets and evaluate it across three test sets: the relatively simple human-involved scenario ("m0"), complex scenarios ("n0"), and real mocap data ("s0"). "Contact" means human contact, "Drop" means object drop.}
 \label{tab:supp_scenario_scaling}
\end{table*}

\begin{table*}[!ht]
\centering
\small
\begin{tabular}{c|cc|cc|cc}
\hline
\multicolumn{1}{c|}{\multirow{2}{*}{}}                                           & \multicolumn{2}{c|}{m0}            & \multicolumn{2}{c|}{n0}                      & \multicolumn{2}{c}{s0}                      \\                                           &  Success          & Contact / Drop / Timeout  &  Success          & Contact / Drop / Timeout  &  Success          & Contact / Drop / Timeout         \\ \hline



w/o FOA & 58.7  & 9.0 / 16.5 / 15.8 & 48.5 & 26.9 / 13.6 / 11.0 & 75.7 & 10.5 / 6.9 / 6.9 \\
w/o FPC & 58.4 & 13.7 / 20.9 / 7.0 & 43.2 & 31.0 / 17.6 / 8.2 & 69.4 & 14.6 / 9.7 / 6.3 \\
Ours & 63.8 & 6.6 / 13.5 / 16.1 & 53.4 & 20.5 / 14.1 / 12.0 & 77.8 & 5.6 / 10.4 / 6.2 \\
\hline




\hline

\hline
\end{tabular}
\caption{\textbf{Evaluation of different demonstration generation strategies.} 
We train various methods on "n0" and evaluate them across three test sets: the relatively simple human-involved scenario ("m0"), complex scenarios ("n0"), and real mocap data ("s0"). "FOA" refers to without future obstacle avoidance, and "FPC" refers to without final pose constraints.}
 \label{tab:supp_safe}
\end{table*}

\subsection{Training Details}
All data generation, training, and evaluation are conducted on an 8-GPU setup with NVIDIA GeForce RTX 3090 GPUs, which ensures high computational efficiency. This setup is user-friendly and widely accessible for researchers with moderate hardware resources. For training, we use a batch size of 512, a learning rate of 0.003, and a weight decay of \(1 \times 10^{-5}\).

\subsection{Experimental Data Details}
\subsubsection{Evaluation on Different Training Scenes}
In the manuscript, we discuss that using the unscalable mocap dataset ``s0'' results in a significant average drop in the success rate by 34.6\%. 
When trained on the simpler human-involved ``m0'' dataset, the policy performs well on the simple testing set, with small drops in success rates of 1.8\% and 1.4\% for ``m0'' and ``s0'', respectively. However, in more complex scenarios, the success rate decreases by 8.1\%. 
This demonstrates that the complexity and diversity of human assets are crucial for the policy’s ability to handle more challenging tasks. 

To provide further insights beyond success rates, we list the detailed failure rates in Table~\ref{tab:supp_scenario_scaling}, categorized into human contact, object drop, and timeout. Among these, \textbf{human contact failures} are the most critical to avoid, as collisions with humans can lead to unsafe outcomes. When training on the simpler mocap dataset ``s0'' compared to ``n0'', the drop rate increases by 17.4\%, 31.0\%, and 19.4\% across the three test datasets. Furthermore, when analyzing the absolute number of object drops, training on ``s0'' results in increases by factors of 3.63, 2.51, and 4.46 compared to training on ``n0''. 

These results indicate that solely relying on unscalable real-world data not only decreases success rates but also significantly amplifies the number of object drops, making the resulting policy highly unreliable. This highlights one of our key insights: relying on a large volume of high-quality simulated data can outperform using a small amount of real-world data. For many HRI tasks, such as handover, the lack of specific task datasets is a major challenge. Therefore, rapidly synthesizing high-quality, large-scale simulated datasets is a promising solution.

Similarly, we observe that training on ``n0'' achieves higher success rates and a lower human contact rate compared to training on ``m0''. 

It is important to note that these metrics are interrelated. For instance, training on ``s0'' often results in the highest timeout rate, as the policy struggles to capture the dynamics of the environment and fails to keep up with human actions. This, in turn, leads to a lower probability of human contact or object drops, as human contact requires approaching the human, and object drops require grasping the object.
However, even under these circumstances, the human contact rate for ``s0'' remains the highest among the datasets, underscoring its limitations. This observation further highlights the challenges of training on unscalable real-world data and emphasizes the need for diverse, high-quality simulated data to improve the policy's generalizability and safety.

\subsubsection{Evaluation on Different Demonstration Safety Concerns}
In the manuscript, we mention that without future obstacle avoidance, the policy’s success rate decreases by an average of 4.0\%, and the human contact rate increases from 10.9\% to 16.1\%. Without final pose constraints, the success rate decreases by an average of 8.0\%, and the contact rate rises to 19.8\%. These results highlight that demonstrations with a simple safety design can significantly reduce human collisions, which are the most critical failures to avoid compared to object drops or timeouts.

As detailed in Table~\ref{tab:supp_safe}, if we analyze the absolute values, we observe that without future obstacle avoidance, human contact increases by approximately one-third, while without final pose constraints, it doubles. These findings further demonstrate the importance of incorporating these safety designs into the demonstrations.

It is worth noting that we do not discuss\textbf{ time} or\textbf{ average success} rates in detail, as these metrics remain largely unaffected across different training assets or demonstration strategies. Average success directly correlates with success rate changes, so our focus remains on \textbf{success rates} and \textbf{human contact rates}, which provide more meaningful insights into the policy’s performance and safety.

\subsection{Discussion on Different Action Types}
In the manuscript, we mention that the model learns 6D control actions (3D translation and 3D rotation) for the robot gripper and 3D control actions (2D translation and 1D rotation) for the base. For the base movement, there is no ambiguity—it corresponds to the egocentric SE(2) transformation of the base. However, for the robotic arm, its motion can be decomposed into the base's movement and the arm's joint positions. Once the base's motion is determined, the 6D action output by the policy is used to solve the inverse kinematics (IK) for the arm's joint angles.

In all our experiments, the 6D action output by the policy represents only the robotic arm’s egocentric action and does not explicitly account for the base's movement. This approach inherently incorporates the base's motion into the action representation. During IK computation, we first calculate the base's future position based on the base action and then compute the arm's joint positions relative to this future position. The advantage of this approach is that the policy does not need to learn the complex relative motion relationship between the base and the arm.

In contrast, when we attempted to learn the robotic arm’s motion relative to the base, the loss during training appeared normal, but the policy failed to infer reasonable actions. This demonstrates that learning the relative motion relationship is highly challenging. We also experimented with incorporating the relative position between the arm's end effector and the base as part of the policy’s input, but this approach proved ineffective. Moreover, this method lacks robustness, as the relative relationship between the base and the arm can vary significantly with different robot models, making zero-cost transfer across robots impractical.


\subsection{More ablations}

\begin{table}[t]
\centering
\setlength{\tabcolsep}{3mm}
\small
\begin{tabular}{cccc}
\hline
\textbf{Methods}              & m0 & n0 &  s0\\ \hline
w/o flow             &   ~~58.0~~ &   ~~47.2~~  & ~~53.5~~ \\
w/o human            &  51.3  &   48.7     & 57.6 \\
w/o coordinated action      & 51.0    & 41.7 & 48.9 \\
deferred feature fusion  & 55.4 & 47.8 & 75.0 \\
only head camera & 50.3 & 43.4 & 50.7 \\
only wrist camera & 43.6 & 38.7 & 41.7 \\
Ours      & \textbf{63.8} & \textbf{53.4} & \textbf{77.8}  \\
\hline
\end{tabular}
\caption{\textbf{Ablations on different policy designs.} We conduct more ablations on various modules, including different fusion strategies and different perception input.
}
 \label{tab:policy_ablation_supp}
\end{table}

We provide additional ablations in Table~\ref{tab:policy_ablation_supp}. In the manuscript, we mention that when decoding the arm and base actions separately using two independent networks (e.g., two PointNet++), we observe a dramatic 17.8\% decrease in success rate. This demonstrates the necessity of decoding both actions simultaneously. 
Furthermore, when using two PointNet++ encoders, concatenating their output global features, and decoding the arm and base actions based on this deferred feature fusion strategy, we observe decreases in success rates of 8.4\%, 5.6\%, and 2.8\% across the test scenarios. These results indicate that performing fusion within the set abstraction layers is a more effective strategy for scene fusion, highlighting the importance of early integration of object and human features in the policy network.

In our experiments, we utilized two cameras to mitigate occlusion issues and improve visual perception. To validate the importance of using multiple cameras, we conducted experiments with a single camera setup. 
When only the head camera is used, the arm frequently occludes the head camera's field of view, resulting in success rate decreases of 13.5\%, 10\%, and 27.1\% across the test scenarios. Conversely, when only the wrist camera is used, there are more instances where objects are not visible (e.g., when the robot raises its arm or the view is obstructed by the robot's body), leading to success rate drops of 20.2\%, 14.7\%, and 36.1\%. 
These results highlight the importance of multiple cameras in ensuring robust visual perception. In a visual policy, it is critical to provide the policy with as much visual input as possible and effectively fuse the information from multiple sources.

\section{Real World Experiments Details}
\label{supp_sec:real_world}
\subsection{Setup}

\begin{table*}[t]
\centering
\small
\begin{tabular}{c|cc|cc}
\hline
\multirow{2}{*}{} & \multicolumn{2}{c|}{Simple setting}                       & \multicolumn{2}{c}{Complex setting}                       \\ \cline{2-5}           
                  & GenH2R\cite{Wang_2024_CVPR} 
                  & Ours          
                  & GenH2R\cite{Wang_2024_CVPR} 
                  & Ours              \\ \hline
1. mini cocoa crisps    & 2 / 5     & 5 / 5     & 2 / 5     & 3 / 5      \\ \hline
2. adhesive tape    & 3 / 5     & 4 / 5     & 2 / 5     & 3 / 5      \\ \hline
 ~~~~3.chewing gum container~~~~  & 1 / 5     & 3 / 5     & 1 / 5     & 3 / 5      \\ \hline
4. instant noodles    & 1 / 5     & 4 / 5     & 1 / 5     & 4 / 5      \\ \hline
5. chicken jerky   & 3 / 5     & 4 / 5     & 1 / 5     & 4 / 5      \\ \hline
6. bottled iced tea   & 2 / 5     & 4 / 5     & 2 / 5     & 2 / 5      \\ \hline
total             & 12 / 30 (40\%) & \textbf{24 / 30 (80\%)} & 9 / 30 (30\%) & \textbf{19 / 30 (63\%)} \\ \hline
\end{tabular}

 \caption{\textbf{User study for sim-to-real experiments.} our method and GenH2R(reprod.) method were evaluated by five individuals for six objects in both the simple and complex settings. Failure scenarios included collisions with the human body, dropping on the ground, or exceeding the time limit ($T_{\text{max}} = 25$ seconds). Our method consistently outperformed the baseline in the real-world handover system in both simple and complex settings, aligning with the results observed in the simulation experiments.
}
 \label{tab:real_supp}
 
\end{table*}

\begin{figure}[htbp]
  \centering
  \includegraphics[width=\columnwidth]{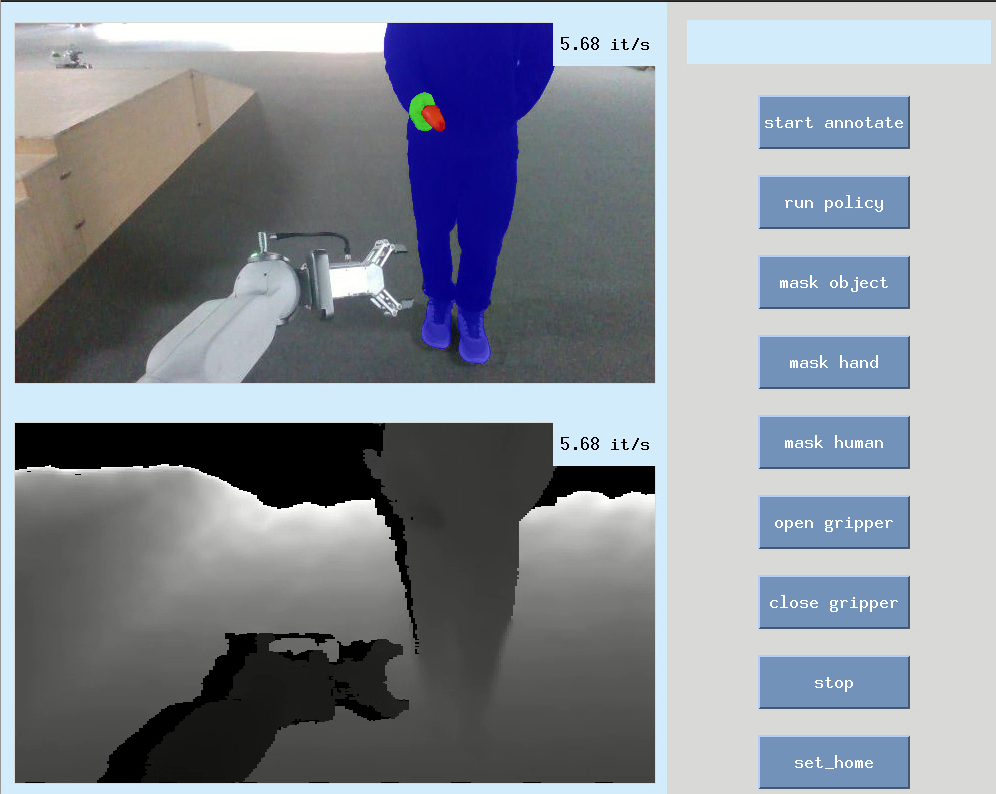}
  \caption{\textbf{Our control panel for real-world experiments.}Powered by SAM2, we can accurately track humans and objects and perform rapid segmentation (6-8 FPS). This enables efficient and precise perception, which is crucial for real-time robotic applications.
}
\label{fig:real_exp_panel}
\end{figure}

Since our policy relies on segmented point clouds of the object, the human hand, and the human body, we use SAM2 \cite{ravi2024sam2} to obtain segmentation masks in our real-world experiments. 

Specifically, we manually annotate the object, the human hand, and the human body in the first RGB image using our control panel as shown in Figure \ref{fig:real_exp_panel}, and then leverage the tracking capability of the SAM2 model to generate segmentation masks in the following RGB images. Combining the segmentation masks and the depth map from the depth camera, we obtain the segmented point clouds to be fed into our policy. After annotating the first RGB image, no further human operation is needed until the end of the experiment.

All of the models used in real-world experiments (including the SAM2 model and our policy model) run on a single NVIDIA GeForce RTX 4090 GPU.

We also provide \textbf{a supplementary video} that showcases extensive real-world demonstrations of our method. The video includes examples from various scenarios, such as healthcare settings and home office environments. It highlights the generalization of our approach across different human heights, diverse objects, and varying obstacle configurations. Additionally, it demonstrates the safety aspects of our method.

\subsection{User Study}

\begin{figure}[t]
  \centering
  \includegraphics[width=\columnwidth]{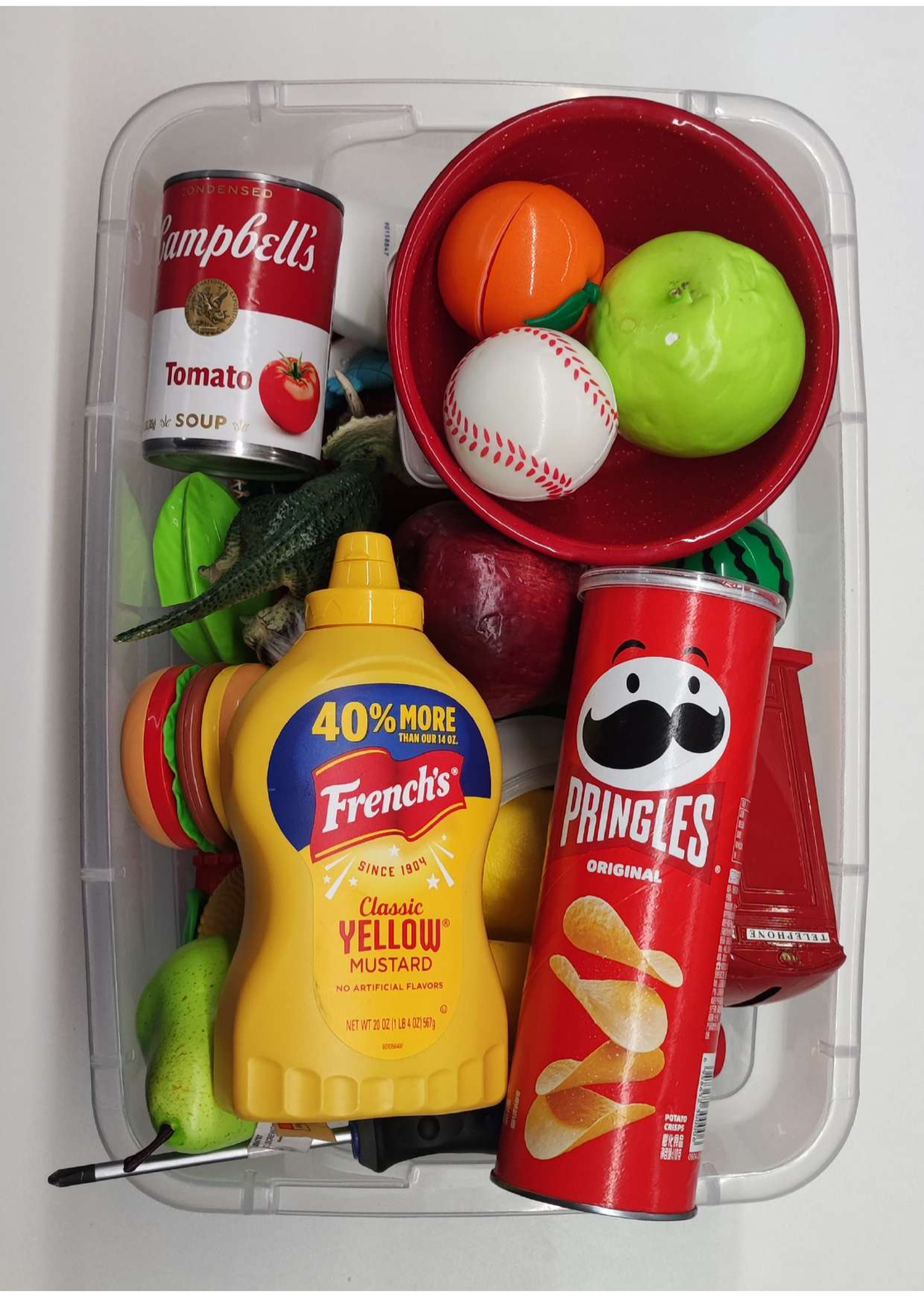}
  \caption{\textbf{Various objects for real-world handover.} The image above displays various objects for mobile handover, including the can, the bowl, the bottles or some plastic objects.}
\label{fig:different_task}
\end{figure}

Table \ref{tab:real_supp} provides a detailed breakdown of the results presented in Table 1 of the manuscript. 
We compare our method with GenH2R(reprod.) across 6 objects in 2 different settings. In the simple setting, users directly hand over the object to mimic ``m0''. In the complex setting, users may sit, go downstairs, or perform adversarial actions to mimic ``n0''. 
Our method is compared with baselines across different settings, revealing a remarkable 40\% improvement in the simple setting and a substantial 33\% improvement in the complex setting from GenH2R.

\section{Limitations and Future Work}
\label{supp_section:limit}

While this paper demonstrates significant progress in the human-to-mobile-robot handover task, we acknowledge certain limitations that could inspire exciting future research directions.

\textbf{Mobile Robot Considerations.}  
In this work, we utilized the Galbot robot, which features a 3-DoF omnidirectional base and a 7-DoF robotic arm. Our policy primarily leverages coordinated base-arm actions. However, with the rapid development of humanoid robots, including bipedal robots and various advanced platforms, different application scenarios may require customized measures to improve policy training, such as modifications to the network architecture. Nevertheless, we believe that our scalable pipeline for generating diverse human motions and the automatic, safe, and imitation-friendly demonstration generation process are broadly applicable. Many of the plug-and-play modules used in our approach can be replaced or adjusted to account for the unique characteristics of other robot types, allowing for the design of tailored policies. We look forward to future work extending mobile handover tasks to a wider variety of mobile robots.

\textbf{Baseline Comparisons.}  
We compared our method with non-end-to-end approaches as well as the state-of-the-art method, GenH2R. According to GenH2R's results, imitation learning significantly outperforms reinforcement learning in simple fixed-based handover tasks. However, in our work, the scenarios are more complex, requiring greater safety considerations, which are challenging to achieve with RL-based methods. As a result, we did not reproduce RL methods in this setting. Of course, there are many potential solutions for mobile human-to-mobile-robot handover tasks, and we hope future frameworks will enable broader comparisons.

\textbf{Real-World Deployment.}  
Our model outputs delta positions, which require robust position control algorithms. Recent advancements in learning velocity or force control present promising directions that our current framework does not yet consider. Additionally, we observed certain challenges during real-world deployment, such as occasional segmentation errors with SAM2 or inaccuracies in depth camera perception. These anomalies pose significant challenges to our model. To address these issues, incorporating random perturbations during training or fine-tuning the model in real-world settings could improve robustness. Some hybrid methods, such as combining synthetic and real-world data for training, could also be effective. While we achieved excellent results using solely synthetic data, we believe further exploration of these methods will enhance sim2real transfer in the future.


\end{document}